\documentclass[letterpaper, 10 pt, conference]{ieeeconf}  % Comment this line out if you need a4paper

\IEEEoverridecommandlockouts                              % This command is only needed if 
\usepackage{times}  %Required
\usepackage{helvet}  %Required
\usepackage{courier}  %Required
\usepackage{url}  %Required
\usepackage{graphicx}  %Required
\usepackage{url}  %Required
\usepackage{amssymb,amsmath,latexsym}
\usepackage{subfigure}
\usepackage{algorithm}
\usepackage[noend]{algpseudocode}
\usepackage{paralist}
\usepackage{url}
\usepackage{xspace}

\newcommand{\vdest}{\ensuremath{v_{\sf dest}}}

\newcommand{\Energy}{\ensuremath{\mathbb{B}}}
\newcommand{\energy}{\ensuremath{\epsilon}}

\newcommand{\Eknown}{\ensuremath{E^{\sf known}}}
\newcommand{\Eunknown}{\ensuremath{E^{\sf unknown}}}

\newcommand{\unknown}{\ensuremath{{\sf unknown}}}

\newcommand{\ShortestPath}{\textsf{\small ShortestPath}\xspace}
\newcommand{\Frontier}{\textsf{\small Frontier}\xspace}

\newtheorem{thm}{Theorem}              % theorem
\newcommand{\commentc}[1]{}
\newcommand{\commentt}[1]{}

\renewcommand{\commentc}[1]{{\bf **Chiu: #1**}}
\renewcommand{\commentt}[1]{{\bf **Ty: #1**}}

\title{\LARGE \bf
Extending the Range of Drone-based Delivery Services by Exploration
}

\author{Tsz-Chiu Au$^{1}$% <-this % stops a space
\thanks{$^{1}$Department of Computer Science and Engineering,
        Ulsan National Institute of Science and Technology, South Korea.
        {\tt\small chiu@unist.ac.kr}}%
}

\begin{document}

\maketitle
\thispagestyle{empty}
\pagestyle{empty}

%%%%%%%%%%%%%%%%%%%%%%%%%%%%%%%%%%%%%%%%%%%%%%%%%%%%%%%%%%%%%%%%%%%%%%%%%%%%%%%%
\begin{abstract}

Drones have a fairly short range due to their limited battery life.  We propose an adaptive exploration techniques to extend the range of drones by taking advantage of physical structures such as tall buildings and trees in urban environments. Our goal is to extend the coverage of a drone delivery service by generating paths for a drone to reach its destination while learning about the energy consumption on each edge on its path in order to optimize its range for future missions. We evaluated the performance of our exploration strategy in finding the set of all reachable destinations in a city, and found that exploring locations near the boundary of the reachable sets according to the current energy model can speed up the learning process.

\end{abstract}

%%%%%%%%%%%%%%%%%%%%%%%%%%%%%%%%%%%%%%%%%%%%%%%%%%%%%%%%%%%%%%%%%%%%%%%%%%%%%%%%
% Contents
%%%%%%%%%%%%%%%%%%%%%%%%%%%%%%%%%%%%%%%%%%%%%%%%%%%%%%%%%%%%%%%%%%%%%%%

\sloppy

\noindent
\section{Introduction}
\label{sec:introduction}

% The recent rise of delivery drones in recent years signifies a new era of
% transportation and logistics.  More and more companies will
% use drones to deliver their packages to their customers.

% The age of drones has arrived, proclaimed by The Economist, an influential
% weekly magazine~\cite{bib:Economist15Drone}.  

A \emph{delivery drone} is an unmanned aerial vehicle for transporting packages.  
Some companies such as Amazon have started to use delivery drones to deliver packages
to their customers.
%\footnote{Amazon Prime Air.~\url{https://www.amazon.com/Amazon-Prime-Air/b?node=8037720011}}
However, drones owned by these companies have a very limited range due to
their short battery life, causing them to be unable to deliver packages beyond a
couple of miles.  The question of how to extend the range of drones is a major
obstacle in the deployment of drone-based delivery in the real world.  Although
some new drone technology have been proposed to extend the range of drones
(e.g., Google X's Project Wing and gas-powered drones), these drones still have
a finite range since they rely on fuel tanks or batteries with limited
capacity.  Some companies have been looking for ways to recharge or swap
batteries (e.g., the automatic landing stations developed by Matternet).  The
drawback of landing the drones for recharging or battery swapping is the
increase of the delivery time.  Hence, how to extend the range of drones on a
single charge remains an important issue.

\begin{figure}[t]
  \centering
  \includegraphics[width=0.7\columnwidth]{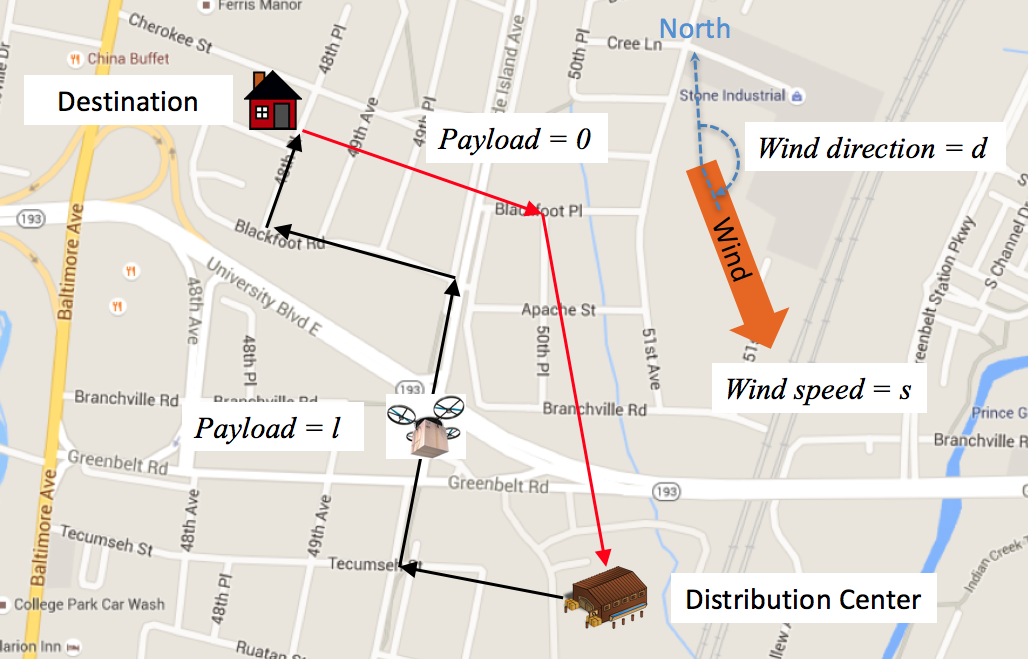}
  \caption{A drone is sent to a destination to deliver a package along a path
(the black arrows) between some buildings so as to avoid flying against a
headwind, and then return to the distribution center (i.e., the base) along a path (the red
arrows) such that it can fly downwind.}
  \label{fig:problem}
  \vspace{-.2in}
\end{figure}

One way to extend the range of drones is to take advantage of physical structures
such as tall buildings and trees in urban environments while taking the wind
direction and wind speed into account.  A drone is a lightweight machine whose
power usage is greatly affected by the airflow, which is influenced by
physical structures in the environment.  A drone flying in
an urban area can take advantage of these structures by flying between
buildings in order to fly downwind as much as possible while avoid flying into the
headwind. As an example, suppose we need to send a drone to deliver a package
to a destination as shown in Figure~\ref{fig:problem}.  Instead of flying
directly to the destination, the drone can save energy by flying between the
buildings (the black arrows) in order to avoid flying against the headwind.  On
its way to fly back to the distribution center, the drone can fly high in order
to take advantage of the downwind (the red arrows). The energy saving can be
translated into an extended range since the drone can now serve a larger area.

Our goal is to extend the \emph{coverage} of delivery service---the set of
destinations that are reachable from a distribution center given a finite
amount of energy.  A coverage is \emph{maximum} if it includes all reachable
destinations.  If we can accurately predict the energy consumption of the
drone, we can easily compute the maximum coverage by using all-pairs shortest
path algorithms.  However, it is difficult to estimate the energy consumption
under the influence of wind and structures in the environment since the
interaction between a drone and an environment can be quite complicated.
Deliberately sending drones to collect data to construct an accurate energy consumption
model for a large area under all kinds of wind conditions
will take too much time that cause downtime to delivery
service.  Moreover, if the environment changes (e.g., a tree falls down or a
new building is constructed), the model may no longer be accurate.  We
therefore propose to learn a model of energy consumption online by exploring
the environment while sending drones to deliver packages.  
In this paper, we propose a new \emph{exploration
strategy} that strategically explores the locations that are currently out of reach.
According to our experiments, our strategy outperforms the one in Nguyen and Au \cite{mybib:Nguyen17extending}
in the task of finding the maximum reachable set in a graph.

\section{Related Work}
\label{sec:related}

Route planning problems for UAVs
can be formulated as Dubins Traveling Salesman Problems (DTSPs)~
\cite{shanmugavel_path_2005} and \cite{ma_receding_2006}, in which 
routes are optimized while satisfying the Dubins vehicle's curvature constraints. As the
optimization for Dubins curves is achieved with respect to distances,
the optimized route plan for a DTSP is only optimal in terms of distance.
However, a trajectory with minimum fuel consumption for other type of UAVs may
differ from the minimum distance trajectory and a minimum distance route plan
may not be fuel optimal~ \cite{zhou_3d_2014}. Therefore, more research on
minimizing energy consumption for battery-powered UAVs is needed.

Previous research often considered the problem of optimizing static soaring
trajectories to minimize energy loss in order to extend the UAV's operating
coverage~\cite{boslough_autonomous_2002,zhao_optimal_2004}. More recent
research addressed the dynamic soaring problem by introducing computational
optimization techniques~\cite{zhao_minimum_2004,sachs_minimum_2005}.  The
increasing popularity of UAVs attracted more research on the soaring
problem. For example, 
\cite{chakrabarty_energy_2009} and \cite{chakrabarty_energy-based_2011} introduced an
energy mapping that indicates the lower bound of necessary energy to reach the
goal from any starting point to any ending point in the map with known wind
information. Exploiting this map can provide a path from a starting point to
a goal with the optimal speed and the heading to fly over each cell in the
path.  An alternative is to use soaring mechanisms to utilize observations to
determine optimal flight~\cite{lissaman_wind_2005,langelaan_gust_2009}. A
further alternative is to define a wind model and utilize on-the-fly
observations to fit this model.  However, there has been little research focusing
on building a mapping from wind condition and payload to energy consumption and
simultaneously using that information to generate paths from a base point to
target points.
\cite{lawrance_autonomous_2011}
presented a Gaussian process regression
approach to estimate the wind map but they devised a reward function to
automatically balance the tradeoff between exploring the wind field and
exploiting the current map to gain energy. 
Moreover, their problem differs from
ours because their problem aims to drive the gliders to predefined locations. In
contrast, our work aims to find \emph{all} locations at which a delivery drone can
arrive (i.e., the maximum coverage) and return to the distribution center with
different payload and wind conditions.  To our knowledge,
no existing work specifically focuses on using learning to find a reachable set
in a graph under energy constraints; instead, most work aims to find the shortest path to a node,
which is different from finding all reachable nodes.
For example, \cite{bib:Dey14Gauss} focuses on building a policy for
an aircraft to get to a single destination while taking advantage of initially
unknown but spatially correlated environmental conditions.
\cite{mybib:Nguyen13} and \cite{mybib:Nguyen14}
presented an energy-constrained planning algorithm for information gathering with an aerial glider.
However, the mapping problem is somewhat different from ours with
different objectives and a lot more physical constraints.
\cite{bib:Yamauchi97frontier} also presented a frontier-based exploration approach which, unlike ours, does not aim to explore the entire map
to find the reachable set.

There exists a rich literature on finding traversal cost between
locations in a map for ground vehicles.  For example, 
\cite{betke_piecemeal_1995} provided a scenario in which a taxi driver learns
about a city map in trips that eventually return to the base. Yet, this work's
objective is to learn the traversal cost of the whole map with a minimum number of edges
traversal. \cite{al_2002_intelligent} proposed to find a
minimum-cost route from a start cell to a destination cell, based
on an assumption that a digitized map
of terrain with connectivity is known.  The route's cost is
defined as the weighted sum of three cost metrics: distance, hazard, and
maneuvering.  However, they do not address the problem of balancing the
tradeoff between exploration and exploitation.  
\cite{sofman_2006_improving}
and 
\cite{mybib:ollis_2007_bayesian}
utilized visual information and a robot's on-board perception system to interpret
the environmental features and learn the mapping from these features to
traversal costs in order to predict traversal costs in other locations where only
visual information is available. Our work, however, does not utilize any
visual information.

Some previous works are concerned with balancing exploration and
exploitation in learning and planning.
\cite{souza_bayesian_2014} dealt with the problem of learning the
roughness of a terrain that the robot travels on.  
\cite{martinez-cantin_bayesian_2009} studied an optimal sensing scenario in which
the robot's objective is to gather sensing information about an environment as
much as possible. They devised Bayesian optimization approaches to dynamically
trade off exploration (minimizing uncertainty in unknown environments) and
exploitation (utilizing the current best solution).  
\cite{bib:Kollar08Trajectory} optimized
trajectories using reinforcement
learning in order address the problem of how a robot should plan to explore an
unknown environment and collect data in order to maximize the accuracy of the
resulting map.  However, the problems in these papers are different from ours
as our objective of learning traversal cost map is to generate optimal energy
paths for drones to extend their range.

% \commentc{It also focuses on building paths for aircraft while taking as much advantage as possible of initially unknown but spatially correlated environmental conditions (winds), proposes a principled way of balancing exploration and exploitation in settings of this kind, and an algorithm for resolving this tradeoff. It still only computes a policy to get to a single destination though, unlike the paper under review.}

% For gas powered unmanned aerial vehicles (UAVs), the problem of
% optimizing their fuel consumption is often seen as optimizing the distance
% because their minimum fuel optimal trajectories are similar to their minimum
% distance optimal trajectories. Accordingly, 

% Our work is somewhat related to real time searching problems that have been
% studied in numerous previous works.  For example, Korf in
% \cite{korf_real-time_1990}, Hern{\'a}ndez et al. in
% \cite{mybib:hernandez_improving_2005}, Fu et al. in \cite{fu_uav_2012} and Rivera et
% al. in \cite{rivera_incorporating_2015} investigated the Learning Real Time A*
% algorithm (LRTA*) and its variants. Davis Furcy et al. \cite{furcy_speeding_2000},
% on the other hand, devised the FALCON method, which speeds up the convergence of
% LRTA*. These studies relate to our work as they start with an unknown map, and
% obtain better paths over traversal repetition. Nevertheless, they make the assumption
% that the agent is equipped with lookahead capability which is not present in our work.  

\section{Energy-Bounded Delivery Drone Problems}
\label{sec:problem}

We consider a scenario in which a company sets up a base (i.e., a distribution center) to
serve a community of households as shown in Figure~\ref{fig:problem}.  The
base has one drone for package delivery to the households.  The
drone, which has a maximum energy $\Energy$, must fly on some
designated trajectories connecting the base to the households.
The trajectories can be segmented into edges such that the trajectories form a
connected, directed graph $G = (V,E)$, where $V = \{ v_0, v_1, \ldots, v_N\}$
is a set of nodes and $E \subseteq V \times V$ is a set of directed edges among
the nodes in $V$.  Let $v_0 \in V$ be the location of the base
and $D \subset V \setminus \{ v_0 \} $ be the set of all nodes corresponding to
the locations of the households.

From time to time, the base receives requests from the
households to deliver packages to their houses.  For our purpose, a \emph{request} is
a pair $(\vdest, l)$, where $\vdest \in D$ is called the \emph{destination}
which is the location of the household and $l$ is the \emph{payload} which is
the weight of the package.  Upon receiving a request $(\vdest, l)$, the
base will first decide whether it is feasible to deliver a
package using a drone, and then send out a drone if it is feasible.
The feasibility depends on whether there exists a cycle $\tau = \rho_1 \oplus
\rho_2$ in $G$, where $\rho_1$ is a path connecting $v_0$ to $\vdest$ and
$\rho_2$ is a path connecting $\vdest$ to $v_0$, such that the drone has enough
energy to fly to $\vdest$ on $\rho_1$ with a payload $l$ and then fly back to
$v_0$ on $\rho_2$ with a zero payload after dropping the package at $\vdest$.
We call the cycle $\tau$ a \emph{trip}.

When a drone traverses an edge $e$ in $G$, it consumes a certain amount of
energy. There are three factors that affect the energy consumption: 1) the
payload $l$, 2) the wind speed $s$, and 3) the wind direction $d$, which is an
angle relative to the north direction.  We say $(l, s, d)$ is a
\emph{configuration}.  We assume that the energy needed to traverse an edge $e$
under a particular configuration $(l, s, d)$ does not change over time. Hence, we use
$\energy(e; l, s, d)$ to denote the energy consumption for traversing an edge
$e$ under $(l, s, d)$.  While our model considers the energy consumption for
traversing an edge, we can actually model the effect of individual trees or
buildings on an edge by replacing the edge with multiple edges, each of them
corresponding to a structure on the edge.  Notice that for any two adjacent
nodes $v_i$ and $v_j$, the energy consumption for flying from $v_i$ to $v_j$
can be very different from flying from $v_j$ and $v_i$, especially if the drone
flies downwind in one direction but flies against a headwind in the other direction.

To simplify our analysis, we assume that the wind speed and the wind direction
do not change during delivery since the flight duration is usually not too long.
However, the wind speed and direction can be different for different requests made
at different time.  Before a drone departs from the base,
we assume we can obtain the current wind speed $s$ and the current
wind direction $d$ from an external information source.
Given a path $\rho$, let $\energy(\rho; l, s, d) = \sum_{e
\in \rho} \energy(e; l, s, d)$ be the total energy a drone consumes when
traversing $\rho$ under $(l, s, d)$.  
Let $\tau = \rho_1 \oplus \rho_2$ be a trip to a destination $\vdest$ such that 
$\rho_1$ is a path connecting $v_0$ to $\vdest$ and $\rho_2$ is a path connecting $\vdest$ to
$v_0$.  The total energy consumption on $\tau$ under $(l,s,d)$ is
$\energy(\tau; l, s, d) = \energy(\rho_1; l, s, d) + \energy(\rho_2; 0, s, d)$.
Then we say a trip $\tau$ to $\vdest$ is \emph{successful} under $(l, s, d)$ if
$\energy(\tau; l, s, d) \leq \Energy$.  

A destination $\vdest \in D$ is \emph{reachable} under $(l, s, d)$ 
if there \emph{exists} a successful trip $\tau$ for $(l, s, d)$.
We also say $\vdest$ is $(l,s,d)$-reachable.
It is easy to check whether $\vdest$ is $(l,s,d)$-reachable if
$\energy(e; l, s, d)$ and $\energy(e; 0, s, d)$ are known for every edge $e$.
First, find the shortest path $\rho_1'$ from $v_0$ to $\vdest$ given
$\energy(\cdot; l, s, d)$. Second, find the shortest path $\rho_2'$ from
$\vdest$ to $v_0$ given $\energy(\cdot ; 0, s, d)$.  Then
$\vdest$ is $(l,s,d)$-reachable \emph{if and only if}
\begin{equation}
  \energy(\rho_1'; l, s, d) + \energy(\rho_2'; 0, s, d) \leq \Energy.
\label{eq:reachable}
\end{equation}
At the beginning, we do not know which destinations satisfy
Equation~\ref{eq:reachable} since we do not necessarily know $\energy(e; l, s,
d)$ and $\energy(e; 0, s, d)$ before exploration, even though
we know the structure of the graph.
However, we assume drones
are equipped with devices that can measure their energy consumption along an edge.  As
we make more and more deliveries, we collect more and more
measurements and hence we will gradually identify more and more destinations
that are $(l,s,d)$-reachable. This knowledge effectively extends the range of a
drone as we can now confidently serve destinations that are further away from
the base.  Ultimately, our goal is to find the set $D_{l,s,d}
\subseteq D$ for all possible configurations $(l,s,d)$ such that all $v \in
D_{l,s,d}$ are $(l,s,d)$-reachable.  We call $D_{l,s,d}$ the \emph{reachable
set} under $(l,s,d)$, which represents the maximum coverage of the base
under $(l,s,d)$.

\section{Energy Models and Current Reachable Sets}
\label{sec:model}

\begin{figure}[t]
  \centering
  \includegraphics[width=0.6\columnwidth]{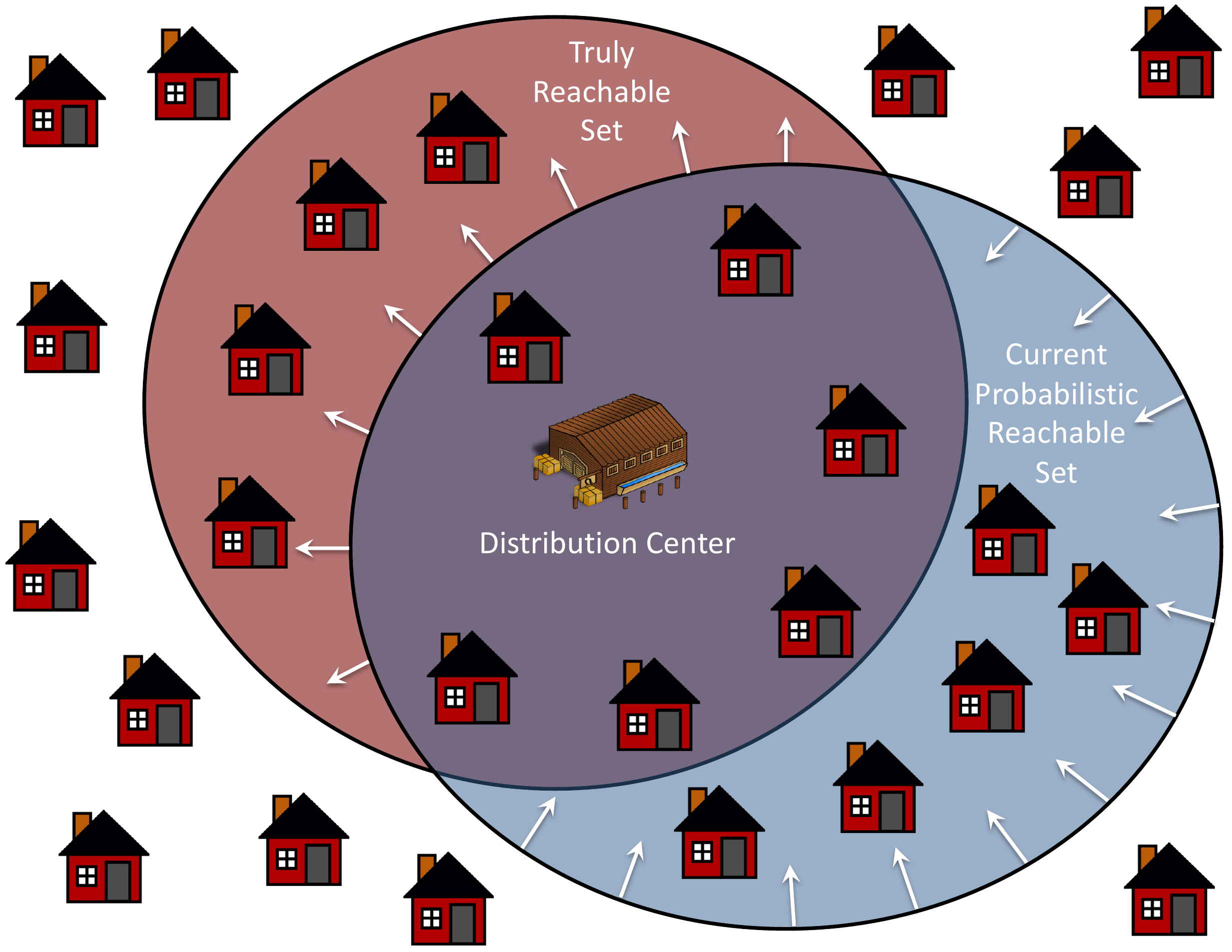}
  \caption{A truly reachable set $D_{l,s,d}$ versus a current probabilistic reachable set
$D_{l,s,d}^t$ at time $t$.  If we assume all unknowns in an energy model follow a normal
distribution, the current probabilistic reachable set is $D_{l,s,d,\phi}^t$ as defined in Section~\ref{sec:model}.}
  \label{fig:reachable_set}
  \vspace{-.2in}
\end{figure}

Given a configuration $(l, s, d)$, we partition the set of edges $E$ into two
sets: $\Eknown_{l, s, d}$ and $\Eunknown_{l, s, d}$, where $\Eknown_{l, s, d}$
is the set of all edges whose energy consumption under $(l, s, d)$ have been
measured by the drone, and $\Eunknown_{l, s, d}$ is the set of all edges whose
energy consumption are currently unknowns.  For each $e \in \Eunknown_{l, s,
d}$, let $u(e; l, s, d)$ be a random variable for $\energy(e; l, s, d)$.  Let
$U_{l,s,d}$ be the set of all random variables for all edges in $\Eunknown_{l,
s, d}$ for a configuration $(l,s,d)$.  To simplify our discussion,
we assume these random variables are independent, though it is possible
to modify our problem formulation to consider covariances among the
random variables.  Suppose the drone maintains an
\emph{energy model} that is used to predict the values of the random variables
in $U_{l,s,d}$.  Let $M^{t}_{l,s,d}$ be an energy model at the current time
$t$. $M^{t}_{l,s,d}$ yields an assignment to the random variables in
$U_{l,s,d}$, which can then be used to compute the \emph{current} reachable set
$D_{l,s,d}^t$ according to Equation~\ref{eq:reachable}.  Clearly, if some
predicted values for some $u(e; l, s, d)$ are different from $\energy(e; l, s,
d)$ which is the true energy consumption, the current reachable set can be
different from the \emph{truly} reachable set $D_{l,s,d}$ based on the true
energy consumption, as shown in Figure~\ref{fig:reachable_set}.  

At the beginning, we are given a prior energy model $M^0_{l,s,d}$, which provides a
rough estimation of the energy consumption.
Upon gathering more measurements, the energy model evolves over time.
Figure~\ref{fig:reachable_set} illustrates the desirable direction of the
evolution: we want the current reachable set eventually become the truly
reachable set by removing false positives from $D_{l,s,d}^t$ and
adding false negatives to $D_{l,s,d}^t$.

Suppose the base sends a drone to deliver a package to $\vdest$
at time $t$, and the drone follows a trip $\tau = \rho_1 \oplus \rho_2$, which
is generated from $M^t_{l,s,d}$ using an \emph{exploration strategy},
where $\rho_1$ is a path connecting $v_0$ to
$\vdest$ and $\rho_2$ is a path connecting $\vdest$ to $v_0$. As the drone
traverses $\tau$, it measures the energy consumption of the set of unknowns
$\unknown(\rho_1)$ and $\unknown(\rho_2)$ on $\rho_1$ and $\rho_2$,
respectively.  The measurement will then be used to update the energy model to
create $M^{t+1}_{l,s,d}$.  Suppose the drone measures $\energy(e; l, s, d)$ for
$u(e; l, s, d) \in \unknown(\rho_1)$. First, we remove $e$ from
$\unknown(\rho_1)$ since $\energy(e; l, s, d)$ is no longer an unknown.
Second, we update the energy consumption of other unknowns using $\energy(e; l,
s, d)$.  One simple way is to use a multivariate linear regression model
trained with the known energy consumption and the characteristics of the
edges to project the energy consumption of the unknowns.
For any edge $e' \neq e$, we estimate $\energy(e'; l', s, d)$ by this equation:
\begin{align}
\hat{\energy}(e'; l', s, d) & = W(l,s,d,L_{e'},\theta_{e'})
\label{eq:regression}
\end{align}
where $L_{e'}$ is the length of $e'$, $\theta$ is the angle between $e$ and the
north direction, and $W$ is a multivariate linear regression model trained with
the edges with known energy consumption, including $\energy(e; l, s, d)$, to
predict the energy consumption of an edge given the parameters $l$,$s$,$d$,
$L_{e'}$, and $\theta_{e'}$.  This estimation is discounted by a weight $w_{e'}
= C_1 \exp(-C_2 d)$, where $d$ is the distance between the middle point of $e$
and the middle point of $e'$, and $C_1$ and $C_2$ are constants.

\section{Exploration Strategies}
\label{sec:algorithm}

Nguyen and Au presented an exploration strategy called \ShortestPath,
which biases the search along with the shortest paths
between the base and the destination using the updated energy
model in the previous trip~\cite{mybib:Nguyen17extending}.
Initially, the strategy is given a prior energy
model $M_{l,s,d}$, which was initialized with a rough estimate of the energy
consumption of all edges under $(l,s,d)$.  In our experiments, the prior energy
model simply sets the energy consumption on an edge to a value proportional to
the length of the value with a small random noise.  For each possible
configuration $(l,s,d)$, there is a $M_{l,s,d}$.  Upon receiving a request to
visit a destination $\vdest$ under $(l,s,d)$, \ShortestPath checks whether a
trip exists to reach $\vdest$.  In addition to $\vdest$ and
$(l,s,d)$, the input parameters of \ShortestPath include the maximum energy
$\Energy$, the starting node $v_0$, and a probability threshold $\phi$.  \ShortestPath
will then use a shortest path algorithm such as Dijkstra's algorithm to compute
the shortest paths to and from the destination according to $M_{l,s,d}$ and
$M_{0,s,d}$.  The shortest paths are combined to form a trip $\tau$.
If the probability of success given
a maximum energy $\Energy$ is greater than or equal to
$\phi$, \ShortestPath will accept the request and return $\phi$;
otherwise, it rejects the request.  If \ShortestPath returns $\phi$,
the energy models will be updated according to the measurements during the traversal of $\tau$.

% \begin{algorithm}[t]
% \caption{An exploration strategy based on the shortest paths
% according to the current energy model.}
% \label{alg:shortest_trip}
% \footnotesize
% \begin{algorithmic}[1]
% \Procedure{\textsc{ShortestPath}}{$\Energy$, $v_0$, $\vdest$, $l$, $s$, $d$, $\phi$, $\alpha$}
%   \State Let $\rho_1 = \textsc{ShortestPath}(v_0, \vdest, M_{l,s,d})$
%   \State Let $\rho_2 = \textsc{ShortestPath}(\vdest, v_0, M_{0,s,d})$
%   \State Let $\tau = \rho_1 \oplus \rho_2$
%   \If { $P(\tau; (1 + \alpha) \Energy, l, s, d) \geq \phi$ }
%     \State \Return $\tau$ \hspace{.2in}
%     \State // Energy models such as $M_{l,s,d}$ and $M_{0,s,d}$ will be
%     \State // updated according to the measurements on $\tau$.
%   \Else
%     \State Reject the request.
%   \EndIf
% \EndProcedure
% \end{algorithmic}
% \end{algorithm}

In this paper, we present another strategy called \Frontier, which improves \ShortestPath by
modifying the trip $\tau$ if $P(\tau; (1+\alpha)\Energy, l, s, d) < \phi$, where $\alpha$ is the \emph{acceptance rate} which is zero unless stated otherwise.
The pseudo-code of \Frontier is shown in Algorithm~\ref{alg:frontier}.
The idea is that $P(\tau;
(1+\alpha)\Energy, l, s, d)$ may be only slightly smaller than $\phi$, which means that
$\vdest$ may actually be quite near the boundary (i.e., the frontier) of the
current reachable set as depicted as the boundary of the blue circle in
Figure~\ref{fig:reachable_set}.  If it is the case, $\vdest$ has a high chance
to be in the truly reachable set even though 
the current reachable set, according to the current energy model, does not
include  $\vdest$.  Therefore, we propose a step called \emph{frontier expansion},
in which \Frontier returns $\tau' = \rho'_1 \oplus \langle v_a, \vdest, v_b \rangle \oplus \rho'_2$,
where $v_a$ and $v_b$ are two randomly-chosen \emph{neighbors} of
$\vdest$ in $G$, $\rho'_1$ is the shortest path from $v_0$ to $v_a$ according
to $M_{l,s,d}$, and $\rho'_2$ is the shortest path from $v_b$ to $v_0$
according to $M_{0,s,d}$. Hence, \Frontier takes the risk
to explore the neighbors of $\vdest$ so as to check whether $\vdest$ is
reachable.  The effect is to expand the frontier of the current reachable set in
the direction of the arrows in Figure~\ref{fig:reachable_set}.
There is a non-zero probability of $\beta$ to perform frontier expansion regardless of
the shortest paths to $\vdest$ (Line~5).
Note that the condition that $P(\rho'_1 \oplus \rho'_2; (1+\alpha)\Energy, l, s,
d) \geq \phi$ at Line~13 is to make sure that $v_a$ and $v_b$ are not
too far away from the current reachable set.
% The risk can be under control if we make
% the new trip is still $\kappa$-safe.

\begin{algorithm}[t]
\caption{The \Frontier exploration strategy.}
\label{alg:frontier}
\small
\begin{algorithmic}[1]
\Procedure{\textsc{Frontier}}{$\Energy$, $v_0$, $\vdest$, $l$, $s$, $d$, $\phi$, $\alpha$, $\beta$}
  \State Let $\rho_1 = \textsc{ShortestPath}(v_0, \vdest, M_{l,s,d})$
  \State Let $\rho_2 = \textsc{ShortestPath}(\vdest, v_0, M_{0,s,d})$
  \State Let $\tau = \rho_1 \oplus \rho_2$ and let $r \in [0,\ 1)$ be a random real number.
  \If { $r \geq \beta$ and $P(\tau; (1+\alpha)\Energy, l, s, d) \geq \phi$ }
    \State \Return $\tau$ \hspace{.2in}
    \State // Energy models such as $M_{l,s,d}$ and $M_{0,s,d}$ are
    \State // updated according to the measurements on $\tau$.
  \Else \hspace{.1in} // frontier expansion
    \State Randomly choose two neighbors of $\vdest$: $v_a$ and $v_b$
    \State Let $\rho'_1 = \textsc{ShortestPath}(v_0, v_a, M_{l,s,d})$
    \State Let $\rho'_2 = \textsc{ShortestPath}(v_b, v_0, M_{0,s,d})$
    \If { $P(\rho'_1 \oplus \rho'_2; (1+\alpha)\Energy, l, s, d) \geq \phi$ }
      \State \Return $\tau' = \rho'_1 \oplus \langle v_a, \vdest, v_b \rangle \oplus \rho'_2$
      \State // Energy models such as $M_{l,s,d}$ and $M_{0,s,d}$ are
      \State // updated according to the measurements on $\tau'$.
    \Else
      \State Reject the request.
    \EndIf
  \EndIf
\EndProcedure
\end{algorithmic}
\end{algorithm}

\subsection{Convergence Analysis}

The fact that \Frontier ignores the energy consumptions of edges $(v_a, \vdest)$ and $(\vdest, v_b)$ at Line~13 in Algorithm~1 guarantees that \Frontier will eventually visit all reachable nodes as long as every node will request delivery indefinitely.

\begin{thm}
\label{thm:converge}
Given a configuration $(l,s,d)$ and $\phi \in [0,\ 1]$, 
the probabilistic reachable set $D^t_{l,s,d,\phi}$ will eventually converge to 
the true reachable set $D_{l,s,d}$ (i.e., there exist a time $t$ such that
$D^t_{l,s,d,\phi} = D_{l,s,d}$),
if every node has a non-zero probability to make a request at every time step.
\end{thm}

Theorem~\ref{thm:converge} provides a guarantee that the true reachable set
can be always found despite of the randomness of the environment and the
exploration process.  Hence, there will be no missing reachable destination
under this exploration strategy.  
The exploration strategy \ShortestPath,  introduced in \cite{mybib:Nguyen17extending},
cannot provide such guarantee. This property is important in practice because we want
to maximize the coverage of the service.  To the best of our knowledge, there is no other
exploration strategy in the literature that can guarantee convergence.

Notice that this theorem can provide such guarantee for
a given configuration $(l,s,d)$ only.  Since $l$, $s$, and $d$ are continuous
values, the drone has to keep track of the energy model of different $(l,s,d)$ in order
to guarantee convergence for any configuration.

\section{Empirical Evaluation}
\label{sec:expr}

We conducted simulation experiments to compare \ShortestPath with \Frontier to see whether extending frontiers can improve the performance.  We implemented a Java simulator which simulates a drone flying in a city.  The simulation uses a physical model of energy consumption to compute the energy consumption when a drone flies along an edge under the influence of wind and buildings. Since we cannot find such model in the literature, we developed an simplified energy model based on the following equations:
\begin{equation}
  \energy(e; l, s, d) = A_c \times (m_0 + l) \times (v_a^2 \times B_c + C_c) \times L \times \frac{D_c}{\cos{\beta}} + E_c
  \label{eqn:energyModel}
\end{equation}
where 
$v_a^2 = (s \times \sin{d})^2 + (V_{max} - s \times \cos{d})^2$,
$\sin{\beta} = (s \times \sin{d} )/ v_a$,
$v_a$ is the velocity of the drone,
$V_{max}$ is the maximum velocity of the drone,
$\beta$ is the angle between the thrust direction and the edge,
$L$ is the length of the edge,
$m_0$ is the weight of the drone without payload, and
$A_c, B_c, C_c, D_c$, and $E_c$ are constants.
This equation provides a rough estimation of the energy consumption under of the effect of wind in the simulation.
Notice that the drone does not have any knowledge about this equation in our experiments.

\begin{figure}[t]
  \centering
  \includegraphics[width=0.95\columnwidth]{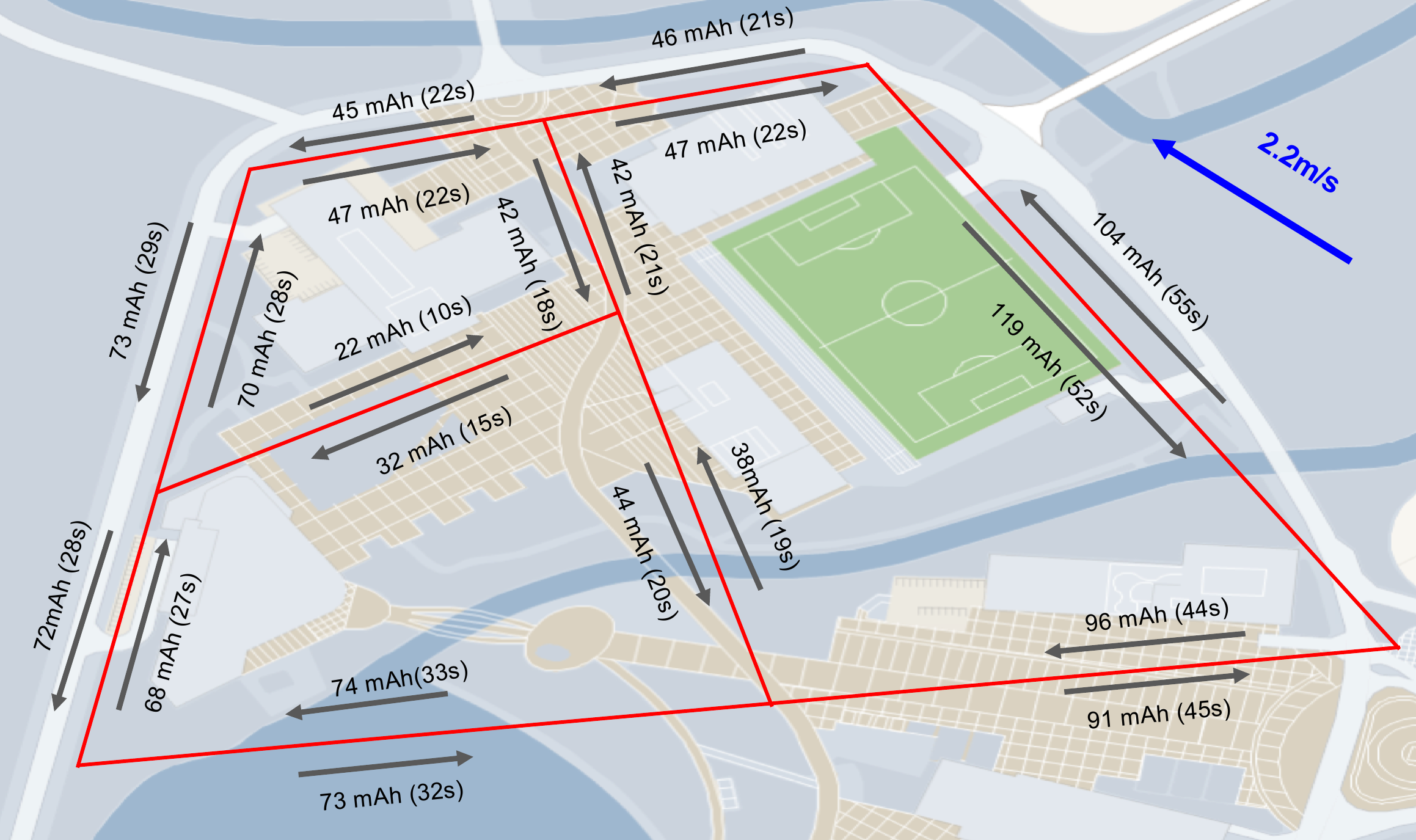}
  \caption{The energy consumption of a real flight. The blue arrow shows the average wind speed and direction. The energy consumption and the flight time are shown on the gray edges. The capacity of the battery is 2200 mAh.}
  \label{fig:energy-data}
  \vspace{-.2in}
\end{figure}

This model is calibrated based on real data. Figure~\ref{fig:energy-data} shows the energy consumption data we collected in a flight.  While the wind speed and the wind direction of the location changes throughout a day, they are quite steady in a short flight duration. Nonetheless, we recorded the average wind speed and the average wind direction. After we collected the energy consumption data of a number of such flights in the real world, we tuned the constants in Equation~\ref{eqn:energyModel} such that the error between the energy consumption predicted by the equation and the real data is minimized.  Note that in our experiments as well as the flight in the real world, we assume drones always fly as fast as possible on an edge. This gives a unique value of energy consumption on all edges.

Three types of maps were considered: random graphs, grids, and road networks in the real world. Ten random graphs were generated by randomly choosing 200 locations as nodes in a 2-D area of size 1000 meters $\times$ 1000 meters, and then connecting each node to five nearest nodes.  Each of the ten grids we generated is a square mesh in which all edges have a length of 100 meters.  A road network is a graph that is constructed from the map data on OpenStreetMap (\url{https://www.openstreetmap.org}). The cities we considered are Berlin, Los Angeles, London, New York City, Seoul, Astana, Sydney, and Dubai. The size of the chosen region of the road networks for the first five cities is 2km $\times$ 2km, whereas the size for the remaining three cities is 5km $\times$ 5km. A total of 28 maps were used in our experiments. In all maps, the node closest to the center of the area is chosen to be the distribution center, and buildings are randomly put along the edges.  Furthermore, we chose the maximum energy $\Energy$ such that approximately 60\% of the nodes are reachable; hence the size of the map does not matter since drones have more energy in larger maps.

\begin{figure*}[t]
  \centering

  \includegraphics[width=0.65\columnwidth]{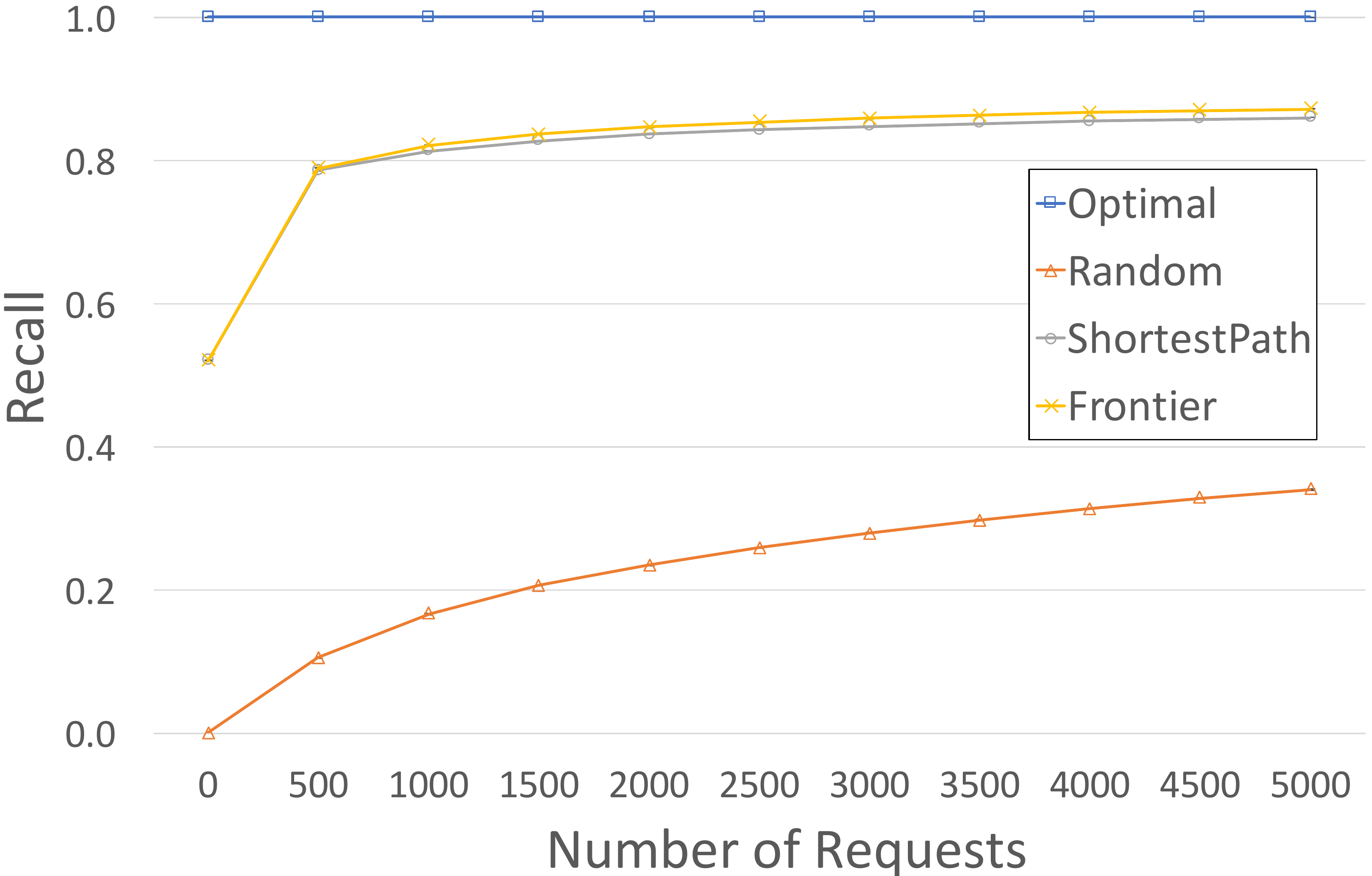}\quad
  \includegraphics[width=0.65\columnwidth]{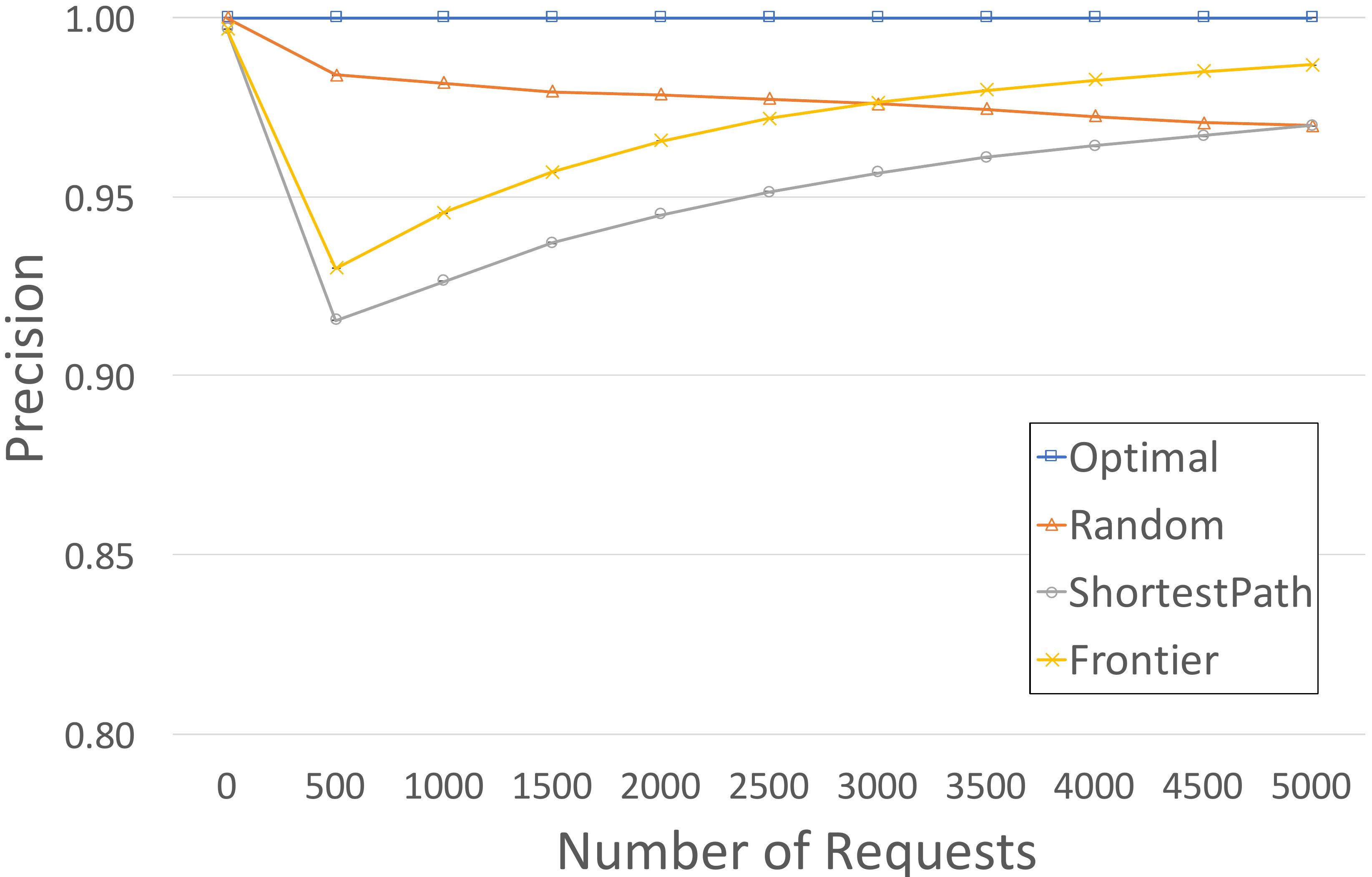}\quad
  \includegraphics[width=0.65\columnwidth]{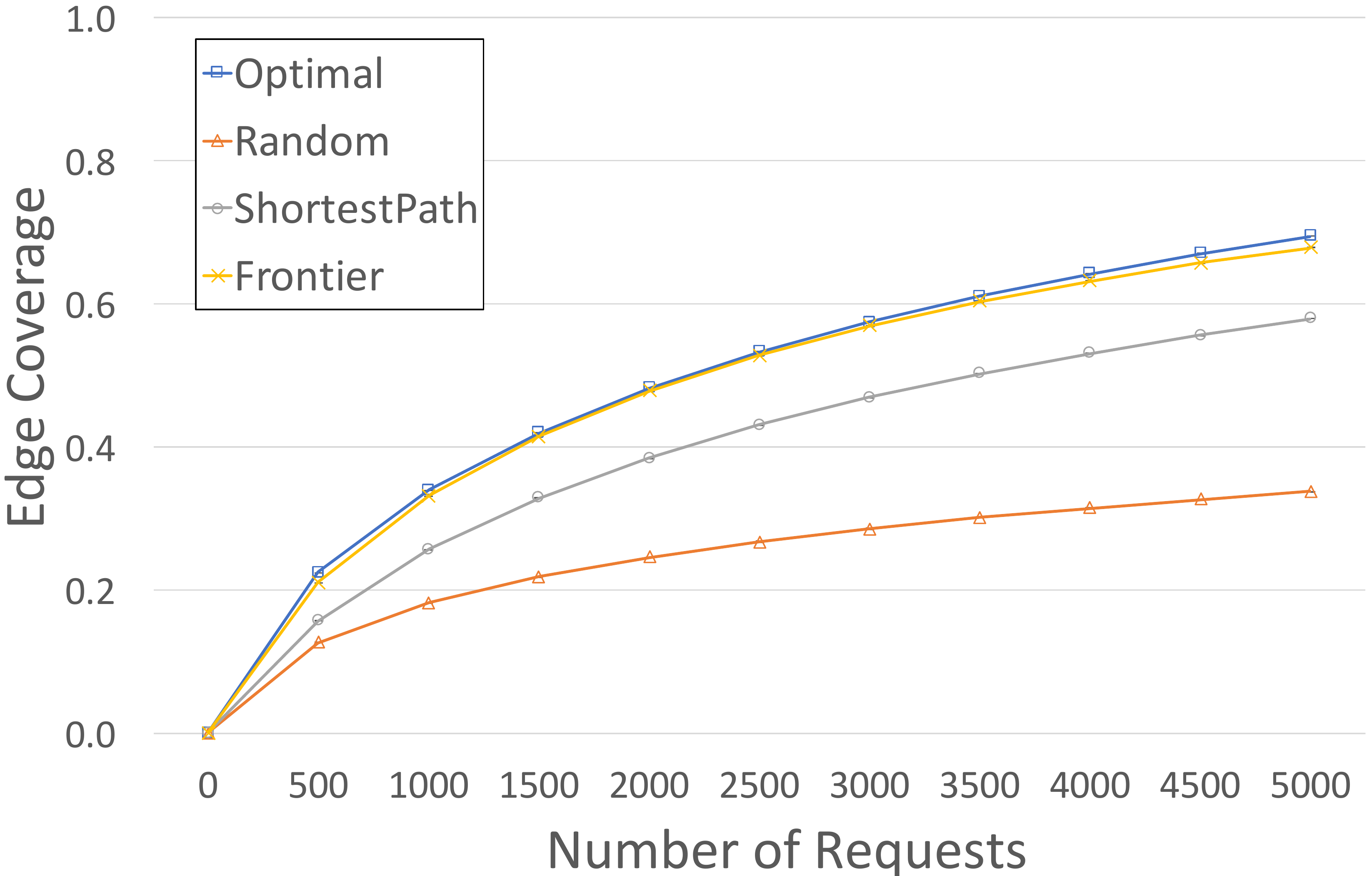}

  % \includegraphics[width=0.75\columnwidth]{figures/expr5000/expr2-recall.pdf}

  % \mbox{}

  % \includegraphics[width=0.75\columnwidth]{figures/expr5000/expr2-precision.pdf}

  % \mbox{}

  % \includegraphics[width=0.75\columnwidth]{figures/expr5000/expr2-edge.pdf}
  
  \caption{The recall, the precision, and the edge coverage as the number of requests increases.}
  \label{fig:expr2}
  \vspace{-.1in}
\end{figure*}

In addition to \ShortestPath with \Frontier, we implemented the optimal
strategy---a strategy that generates trips based on full knowledge of the true
energy consumption of all edges---and a random strategy---a strategy that
randomly generates a trip to the destination--as baselines.  Given a map and an
exploration strategy, the simulator proceeded by generating a sequence of
$5000$ delivery requests with \emph{random configurations} (i.e.,
the wind speeds, directions, and payloads can be different in different requests)
at random nodes, including both reachable and unreachable.  
The payloads, wind speeds and wind directions
in these configurations were drawn randomly from a uniform distribution.
Internally, the drone discretized configurations into 100 different configurations (5 values for wind speed, 5 values for wind direction, and 4 values for payload), and
it learns 100 energy models of these configurations in parallel.
The contribution of a measurement of the energy consumption of an edge
to every edge in every model is determined by Equation~\ref{eq:regression}
as described in Section~\ref{sec:model}. \Frontier will
only use one of the energy models whose configuration is the closest to a given configuration.

At the beginning, the prior energy models
in \ShortestPath and \Frontier are initialized by setting the energy consumption
of an edge to be a random number in $[0.5 k L,\ 1.5 k L] $, where
$L$ is the length of the edge and $k$ is a constant.
Upon receiving a request, a strategy can decide
whether to accept the request. After each flight, the simulator records whether
the delivery is successful and asks the strategy for the current probabilistic
reachable set $D_{l,s,d,\phi}$, where $\phi$ is 0.95.
The probability of frontier expansion $\beta$ is 0.05.
Then we compare $D_{l,s,d,\phi}$ with the
truly reachable set.  The main criteria of the comparison are
\emph{recall}---how many nodes in the truly reachable set are present in the
probabilistic reachable set---and \emph{precision}---how many nodes in the
probabilistic reachable set are truly reachable.  We also compare their
\emph{edge coverage}---how many visited edges are on the shortest paths to any
truly reachable destinations.  Note that we do not report the number of
unsuccessful delivery separately because these unsuccessful delivery will lower the
learning speed, and thus their negative effects have been reflected in the learning rate.
All drones return to the base safely in our experiments.

We conducted an experiment to compare the exploration strategies as
the number of requests increases (see Figure~\ref{fig:expr2}).  Notice that
each data point in these figures is an average of 600 numbers.  The 95\%
confidence intervals are shown as the tiny error bars at the data points in
these figures. According to Figure~\ref{fig:expr2}, the precision and the edge
coverage of \Frontier were much better than \ShortestPath as shown in
Figure~\ref{fig:expr2}, while the recalls were almost the same.  Moreover, the
edge coverage of \Frontier seems to be quite close to the edge coverage of the
optimal strategy, which means that the set of edges visited by \Frontier and
the optimal strategy are nearly the same.  However, we anticipate that as the
number of requests increases, the edge coverage of \Frontier cannot reach 100\%
as the optimal strategy does, because there are some edges visited by the
optimal strategy that would be difficult for \Frontier to explore.
However, if we implement the random exploration scheme as described 
at the end of the last section, \Frontier will eventually explore all edges
to all reachable nodes in a graph, and hence give the same result as
the optimal strategy in the long run.

%%%%%%%%%%%%%%%%%%%%%%%%%%%%%%%%%%%%%%%%%%%%%%%%%%%%%%%%%%%%%%%%%%%%%%%%%%%%

Our second experiment focused on the effect of virtually increasing the 
maximum energy of a drone when deciding whether a trip is acceptable.
Let $\alpha$ be the percentage of the increase of the maximum energy
as defined previously. A large value of $\alpha$ will allow the
drone to accept more requests, meaning that there will be more ``risky''
exploration---flying to destinations even though the current energy model
suggests that there is not enough energy to fly to these destinations.
The experiment setting was the same as the setting in the first experiment
except that the number of requests was reduced to $2000$.  Apart from recall
and precision, we also report the acceptance rate of requests (i.e., how many
requests have been accepted for delivery), the success rate of requests (i.e.,
among all accepted requests, how many delivery have been successful), and the
delivery rate (i.e., among all requests, how many delivery have been
successful). Notice that the delivery rate is equal to the acceptance 
rate times the success rate of requests.

Table~\ref{tb:threshold} shows the results after $2000$ requests.  Each number
in Table~\ref{tb:threshold} is an average of 600 numbers.  As can be seen, all
statistics except the success rate and the recall increases as $\alpha$
increases.  It is because more exploration were encouraged as the drone thinks
it has more energy than it actually has, and that helps the drone to gather
more accurate information about the actual energy consumption to increase the
precision.  The success rate of delivery decreases because some of these
exploration will fail due to the lack of energy. However, the delivery rate
only increases little even after we accept more requests. This means that many
of the new risky exploration end up in failure.  Nonetheless, a larger value of
$\alpha$ does have a positive effect on precision and the delivery rate, hence
if we do not mind many delivery will fail, we should set $\alpha$ to a larger
value.  But the downside is that the recall slightly decreases as the strategy
behave more like a random strategy whose recall is low.

% From Table~\ref{tb:threshold} we can infer that the reason why \Frontier
% performs better than \ShortestPath in the first experiment: \Frontier
% accepts more requests even when $\alpha = 0$. But the effect can be compensated
% by increasing $\alpha$ to $20\%$, at which \ShortestPath performs roughly the
% same as \Frontier at $\alpha = 0$, despite that \Frontier still has a
% higher precision.  Hence, extending frontiers in \Frontier is a better choice than
% increasing $\alpha$ alone if we want to encourage more exploration because 
% extending frontier provides more accurate information about which exploration
% is promising.

\begin{table}
\centering
\caption{The acceptance rate, the success rate, the delivery rate, the
recall, and the precision of the two exploration strategies as the maximum
energy increases.  The 95\% confidence intervals of all numbers are less than
0.001.}
\label{tb:threshold}
\footnotesize
\vspace{-.05in}
\begin{tabular}{|c|c|c|c|c|c|}
\hline
& \multicolumn{5}{c|}{\ShortestPath} \\
\cline{2-6}
  $\alpha$ & Accept & Success & Delivery & Recall & Precision \\
\hline
 0\% &  0.534 & 0.857 & 0.458 & 0.836 & 0.946 \\
 10\% & 0.645 & 0.752 & 0.458 & 0.849 & 0.958 \\
 20\% & 0.755 & 0.654 & 0.494 & 0.850 & 0.969 \\
 30\% & 0.843 & 0.585 & 0.493 & 0.845 & 0.976 \\
 40\% & 0.909 & 0.542 & 0.492 & 0.838 & 0.981 \\
 50\% & 0.950 & 0.518 & 0.492 & 0.834 & 0.983 \\
\hline
\end{tabular}
\vspace{.1in}
 
\begin{tabular}{|c|c|c|c|c|c|}
\hline
& \multicolumn{5}{c|}{\Frontier} \\
\cline{2-6}
  $\alpha$ & Accept & Success & Delivery & Recall & Precision \\
\hline
 0\% &  0.724 & 0.658 & 0.477 & 0.846 & 0.965 \\
 10\% & 0.965 & 0.588 & 0.487 & 0.842 & 0.974 \\
 20\% & 0.903 & 0.543 & 0.490  & 0.836 & 0.980 \\
 30\% & 0.950 & 0.518 & 0.492 & 0.831 & 0.983 \\
 40\% & 0.975 & 0.504 & 0.491 & 0.829 & 0.984 \\
 50\% & 0.989 & 0.498 & 0.492 & 0.827 & 0.985 \\
 \hline
\end{tabular}
\vspace{-.1in}
\end{table}

% The symbols in these equations are defined in Table~\ref{table:energy_symbol}.

% \begin{table}[t]
%   \centering
%   \caption{The definition of the symbols in Equation~\ref{eqn:energyModel}.}
%   \label{table:energy_symbol}
%   \tiny
%   \begin{tabular}{lcl}
%     \hline
%     $\energy(e; l, s, d)$ & $=$ & Energy consumption on edge $e$ under a configuration $(l, s, d)$ \\
%     $v_a$ & $=$ & The velocity of the drone \\
%     $V_{max}$ & $=$ & The maximum velocity of the drone ($m/s$)  \\
%     $\beta$ & $=$ & The angle between the thrust direction and the edge direction \\
%     $(l,s,d)$ & $=$ & The configuration \\
%     $L$ & $=$ & The length of the edge \\
%     $m_0$ & $=$ & The weight of the drone without payload \\
%     $A_c, B_c, C_c, D_c, E_c$ & $=$ & Constants \\
%     \hline
%   \end{tabular}
% \end{table}

%%% Local Variables:
%%% mode: latex
%%% TeX-master: "main"
%%% End:

\section{Conclusions and Future Work}
\label{sec:conclusion}

We addressed the drone exploration problem put forth by \cite{mybib:Nguyen17extending} and proposed a new exploration strategy called \Frontier, which encourages exploration near the frontier to speed up the learning process.  Unlike the \ShortestPath strategy in \cite{mybib:Nguyen17extending}, \Frontier guarantees to converge to the truly reachable set.  While this paper focuses mainly on solving a real-world problem for delivery drones, the underlying problem of finding all reachable nodes in an unknown graph is theoretically interesting by itself since previous works mostly focus on finding one reachable node only in an unknown graph.  While we considered flying one drone at a time, our framework allows multiple drones to fly at the same time. If there are $N$ drones flying in parallel all the time, we expect the learning rate can speed up by a factor of $N$. In the future, we intend to leverage the shared knowledge among multiple drones to speed up the learning process.

\section*{Acknowledgments}
This work has been taken place in the ART Lab at Ulsan National Institute of Science \& Technology. ART research is supported by NRF (2.190315.01 and 2.180557.01).

\bibliographystyle{IEEEtran}
\bibliography{bib/auto,bib/chiu,bib/ref}

\end{document}